\documentclass{article}
\usepackage{spconf,amsmath,graphicx}
\usepackage{hhline}
\usepackage{bm}
\usepackage{bbm} 
\usepackage{tabularx}
\usepackage{multirow}
\usepackage{booktabs} 
\usepackage{mathrsfs}
\usepackage{tabulary}
\usepackage{makecell}
\usepackage{amsfonts,amssymb}
\usepackage[ruled,linesnumbered]{algorithm2e}  
\usepackage{tabularx}
\usepackage{color, colortbl}
\definecolor{Gray}{gray}{0.8}
\definecolor{LightGray}{gray}{0.8}

\title{
Parallel Sentence-Level Explanation Generation for \\ Real-World Low-Resource Scenarios
}

\name{Yan Liu$^1$, Xiaokang Chen$^2$, Qi Dai$^1$}
\address{$^1$Microsoft Research Asia \\ $^2$School of Intelligence Science and Technology, Peking University}
\begin{document}
\maketitle

\begin{abstract}
In order to reveal the rationale behind model predictions, many works have exploited providing explanations in various forms.
Recently, to further guarantee readability, more and more works turn to generate sentence-level human language explanations.
However, current works pursuing sentence-level explanations rely heavily on annotated training data, which limits the development of interpretability to only a few tasks.
As far as we know, this paper is the first to explore this problem smoothly from weak-supervised learning to unsupervised learning.
Besides, we also notice the high latency of autoregressive sentence-level explanation generation, which leads to asynchronous interpretability after prediction.
Therefore, we propose a non-autoregressive interpretable model to facilitate parallel explanation generation and simultaneous prediction.
Through extensive experiments on Natural Language Inference task and Spouse Prediction task, we find that users are able to train classifiers with comparable performance $10-15\times$ faster with parallel explanation generation using only a few or no annotated training data.
\end{abstract}


\section{Introduction}
\label{sec:intro}
Recently, deep learning has developed rapidly \cite{chen2022context,chen2022d,liu2021enhance,chen2022group,tang2022not,chen2022conditional,meng2021conditional,chen2021semi,chen2020bi,chen20203d,chen2020real,tang2022compressible,tang2022point}. 
The interpretability of black-box neural networks has aroused much attention and the importance of interpreting model predictions has been widely acknowledged.
Previous interpretation works provide explanations in various forms as the rationale lying behind model decisions, such as attention distribution \cite{xu2015show}, heatmap \cite{samek2017explainable}, input keywords \cite{thorne2019generating}, etc.
Due to the better human readability, many works exploit generating sentence-level human language explanations to better interpret model predictions and have achieved promising performance \cite{camburu2018snli,2022mpii}.

However, sentence-level explanations are hard to achieve in real-world scenarios due to the high latency of autoregressive explanation generation and the severe reliance on human-annotated explanations. For instance, e-INFERSENT\cite{camburu2018snli} autoregressively generates every explanation token, leading to much higher inference latency.
In comparison, although some explanations lack readability\cite{gururangan2018annotation}, such as the attention-based heatmap explanation and post-hoc alignment map explanation, these explanations can be generated almost simultaneously with predictions. 
Moreover, in spite of readability, previous works that generate sentence-level explanations rely heavily on numerous human-annotated explanations during training. Nevertheless, datasets containing human-annotated explanations are rare due to the high cost.

To alleviate these problems, in this work, we introduce the Classification Non-Autoregressive Transformer (C-NAT) framework for simultaneous classification and parallel sentence-level explanation generation with weakly-supervised and unsupervised learning strategies.
To accelerate the explanation generation, we adopt the architecture of the non-autoregressive generation model NAT \cite{gu2017non} to generate all tokens in parallel. 
We also equip the non-autoregressive generation model with a label predictor for simultaneous label prediction.
Besides, to better accommodate real-world low-resource scenarios, we propose our weakly-supervised learning and unsupervised learning strategies.
Specifically, inspired by \cite{hancock2018training}, we first extract a set of labeling functions and the corresponding explanation templates from a small number of human-annotated samples, and then use these labeling functions and explanation templates to produce pseudo labels and explanations for a large amount of unlabeled data.
For the unsupervised learning scenario, we utilize the back-translation mechanism to paraphrase the input sequences as the pseudo explanation targets, and apply a pre-trained language model to refine the predicted explanations during training.
\begin{figure*}[t]
\vskip -0.15in
	\begin{center}
		\includegraphics[width=0.6\linewidth]{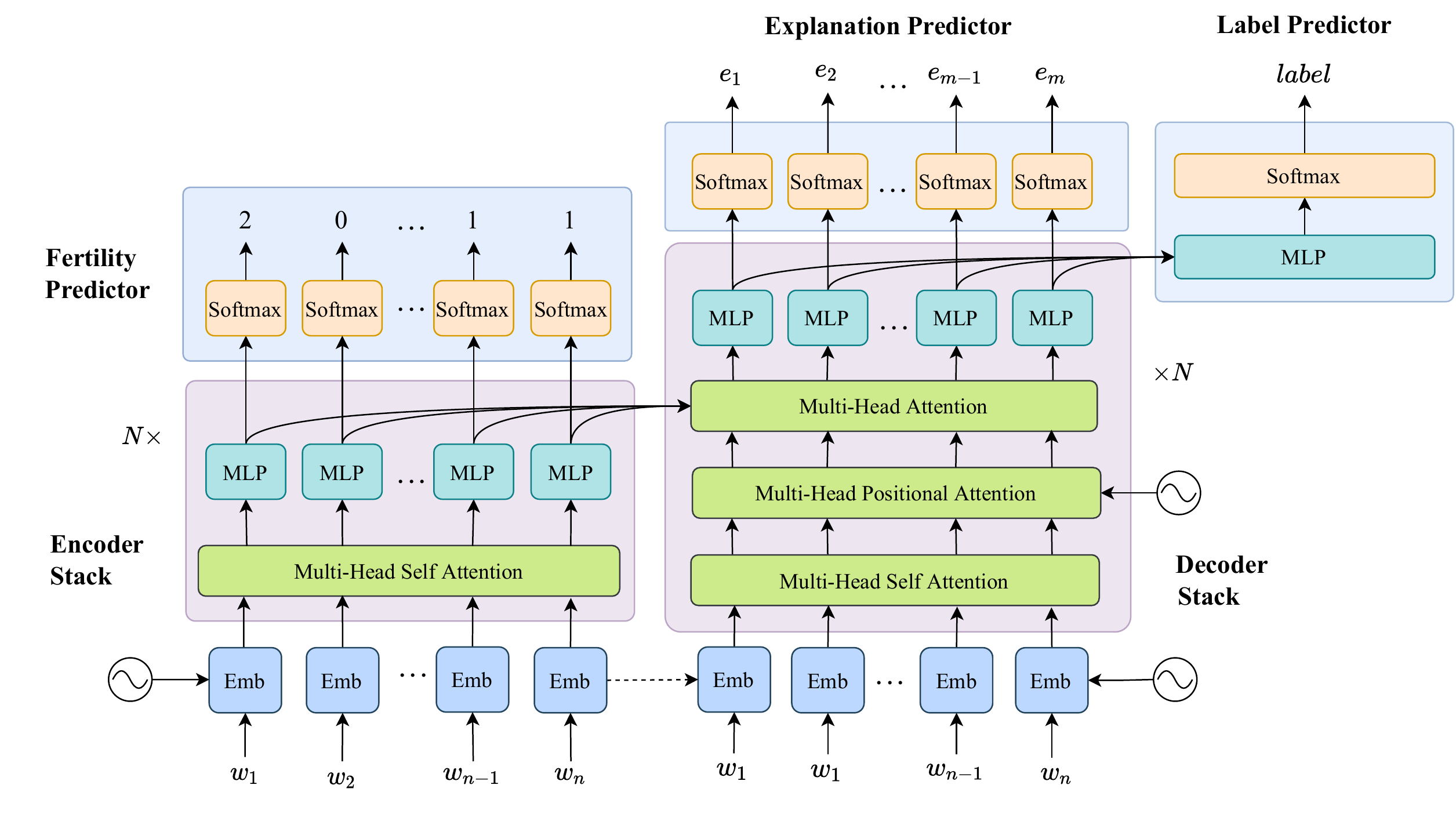}
		\caption{\label{fig1} The overall architecture of our C-NAT.
		}
	\end{center}
	\vskip -0.15in
\end{figure*}
We verify the effectiveness of our approach on the Natural Language Inference (NLI) task and the Spouse Prediction (SP) task. 
Main contributions of this work are three-fold:
\begin{itemize}
\item We propose novel weakly supervised learning and unsupervised learning strategies to accommodate interpretable models to real-world low-resource scenarios.
\item We introduce our C-NAT to support parallel explanation generation and simultaneous prediction.
We also propose to leverage a pre-trained language model as a discriminator to generate more fluent explanations. 
\item 
Experimental results show that our C-NAT can generate parallel fluent explanations and improve classification performance with significant inference speedup, even with few or no human annotations.
\end{itemize}

\section{Model Architecture}
In this section, we introduce the architecture of C-NAT, which modifies the non-autoregressive generation model NAT \cite{gu2017non} to support simultaneous label prediction and parallel sentence-level explanation generation.
As shown in figure \ref{fig1}, C-NAT consists of the following five modules: an encoder stack, a decoder stack, a fertility predictor, an explanation predictor for parallel explanation tokens generation, and a label predictor for simultaneous label prediction.

\subsection{Encoder and Decoder}
We adopt the Transformer\cite{vaswani2017attention} as the backbone. 
To enable non-autoregressive interpretation, following \cite{gu2017non}, the decoder is modified in three aspects: input sequence, self-attention mask, and positional encoding.
For input sequence modification, because previously generated tokens are unavailable under the non-autoregressive setting, we use a fertility predictor first to predict the length of the target explanation and produce decoder input with the tokens copied from the encoder input.
For the modification of the self-attention mask, because the decoder input is the copied sequence of encoder input, the self-attention module is allowed to attend all positions, rather than only left positions in the conventional Transformer decoder. 
Therefore, the self-attention mask is replaced with a non-causal mask in our non-autoregressive decoder. 
For positional encoding modification,
different from the self-attention module, the positional attention module uses positional encoding as the query and key, and the hidden representations from the previous layer as the value.

\subsection{Fertility predictor}
To generate the decoder input sequence for non-autoregressive interpretation, we copy and repeat the tokens from the encoder input. 
The fertility predictor is used to predict the number of times each token is copied, referred to as the \textit{fertility} of each corresponding token \cite{gu2017non}.
Specifically, given the input sentence of the encoder $X=\{x_1, x_2, ..., x_S\}$, the fertility predictor is fed with the encoded feature $H=\{h_1, h_2, ..., h_S\}$, and generates the fertility sequence $F = \{f_1, f_2, ..., f_S\}$.
Finally, the input sequence of the non-autoregressive decoder is $Y=\{y_1, y_2, ..., y_T\}=\{\{x_1\}_{i=1}^{f_1}, \\ \{x_2\}_{i=1}^{f_2}, ..., \{y_S\}_{i=1}^{f_S}\}$ with length $T = f_1 + f_2 + ... + f_S$, where $\{y_s\}_{i=1}^{f_s}$ denotes the token $x_s$ is repeated for $f_s$ times.

\subsection{Explanation Predictor and Label Predictor}
The explanation predictor and label predictor are used to generating each token of the explanation sentence and classification label simultaneously.
Given the output hidden states of the decoder stack $H^d=\{h^d_1, h^d_2, ..., h^d_T\}$, each explanation token $e_t$ is generated with the probability $p_E(e_t) = \text{Softmax}(h^d_t)$, and the explanation sentence $E=\{e_1, e_2, ..., e_T\}$ is generated in parallel with the probability $p_E(E|X;\theta) = \prod_{t=1}^T p_E(e_t|X;\theta)$.
Meanwhile, the label predictor projects the hidden states with an MLP layer and the mean pooling operation, resulting in the label prediction $L$ with the probability $p_L(L|X;\theta)$.

\section{Training Strategy}

\subsection{Fully-supervised Learning}
\label{ssec:fully}
In the explanation available scenario, the fully-supervised training objective function of our model is the combination of the label prediction loss, the explanation prediction loss, and the fertility prediction loss.
Besides, we also apply the pre-trained language model as an extra constraint on the objective function to encourage generating explanations of more fluency and diversity. 
Then the pre-trained language model with parameters $\theta_{LM}$ estimates the log-likelihood of each predicted explanation sentence $E'$  as $\log p_{LM}(E';\theta_{LM})$. To enable the gradient backpropagation  from the pre-trained language model to the C-NAT model, the product of the predicted probability distribution $p_E(e_t|X;\theta)$ and the word embedding vectors is used as the input embedding of the explanation token $e_t'$ in the pre-trained language model. The additional loss term $\mathcal{L}_{LM}$ is adopted to optimize the explanation generation by maximizing the estimated log-likelihood of the pre-trained language model over the training dataset.
Finally, the fully-supervised training objective function of our C-NAT model is formulated as: 
\begin{equation}
\resizebox{0.6\hsize}{!}{$
\label{eq_full_loss}
\mathcal{L} =  \mathcal{L}_L + \lambda_E \mathcal{L}_E + \lambda_F \mathcal{L}_F + \lambda_{LM} \mathcal{L}_{LM}
$}
\end{equation}
, where $\lambda_L$, $\lambda_E$, $\lambda_F$ and $\lambda_{LM}$ are  hyperparameters for each loss term.

\begin{table}[t]
\resizebox{0.47\textwidth}{!}{
	\centering
	\small
	\begin{tabular}{l c c c c}
		\toprule
		\textbf{Datasets}
		& \textbf{Train}
		& \textbf{Val}
		& \textbf{Test}
		& \textbf{Annotated/Total}\\ 
		\hline
		e-SNLI & 549367 & 9842 & 9824 & 570K/570K \\
		SP-Pseudo & 22195 & 2796 & 2796 & 30/22195 \\
		SNLI-Pseudo & 549367 & 9842 & 9824 & 0/570K \\
		\bottomrule
	\end{tabular}}
\caption{Statistics of datasets. 
}
\label{data_stt}
\vskip -0.05in
\end{table}

\begin{table*}[tb]
	\centering
	\small
	\vskip 0.15in
\resizebox{0.85\textwidth}{!}{
	\begin{tabular}{l c c c c c c c c c}
		\toprule
		\textbf{Methods}
		& \textbf{BLEU}$^{\uparrow}$ & \textbf{Rationality}$^{\uparrow}$ 
		& \textbf{PPL}$^{\downarrow}$ & \textbf{Inter-Rep}$^{\downarrow}$	& \textbf{NE-Acc}$^{\uparrow}$	& \textbf{Acc}$^{\uparrow}$
		& \textbf{Latency}$^{\downarrow}$ & \textbf{Speedup}$^{\uparrow}$
		\\ \hline
		Dataset$^{\dagger}$ &  22.51 & 100.00 & 30.00 & 0.40 & 100.00 & 100.00 & - & - \\
		Transformer(AT) &  20.33 & 80.16 & 27.04 & 0.51 & 80.62 & 79.46 & 793ms & $1.27\times$ \\
		e-INFERSENT(AT) &  \textbf{22.40} & 84.79 & \textbf{10.58} & 0.72 & \textbf{84.01} & 83.96 & 1006ms & $1.00\times$ \\
		\hline
		C-NAT &	21.19 &	 \textbf{85.10} & 34.71 & \textbf{0.30} & 82.41 & \textbf{85.23} & \textbf{47ms} & $\mathbf{21.40\times}$ \\
		\ \ \ w/o LM  & 20.87 & 84.51 & 46.77 & 0.32 & 82.41 & 85.19 & 47ms & $21.40\times$ \\
		\ \ \ w/o NAR & 19.33 & 82.06 & 61.14 & 0.47 & 82.41 & 80.47 & 734ms & $1.37\times$ \\
		\ \ \ w/o LCE & 21.04 & 84.16 & 36.21 & 0.33 & 36.68 & 39.31 & 47ms & $21.40\times$ \\
		\bottomrule
	\end{tabular}}
\caption{ Automatic evaluation results in NLI task with full-supervision. The higher$^{\uparrow}$(or smaller$^{\downarrow}$), the better.
$^{\dagger}$We evaluate the ground truth with our metrics.
Latency is computed as the time to decode a single output sequence without mini batching, averaged over the whole test set.
At the bottom, we present the results of the ablation study. 
		}
\label{result_nli_overall}
\end{table*}

\subsection{Weakly-supervised Learning}
\label{ssec:weak}
In the more practical scenario, where only a few human annotated explanations are available, we introduce the weakly-supervised learning strategy to generate the pseudo explanations and pseudo labels for the large-scale unlabeled data. 
Firstly, we extract the labeling functions along with the explanation templates from a small number of human-annotated samples.
Then, we use the labeling functions and the explanation templates to annotate the pseudo labels and explanations for the large-scale unlabeled data.
Due to the wide divergence in accuracy and coverage of the labeling functions, the data programming method \cite{ratner2016data} is applied for label aggregation \cite{hancock2018training}, where a learnable accuracy weight $w_m$ is assigned to each labeling function $f_m(\cdot)$, and the final pseudo label is selected as the label with the largest aggregated accuracy weight.
As for the labeling function with the highest contribution to the pseudo label,
we select the corresponding explanation template $E_m^\text{temp}$ and generate the pseudo natural language explanation $\{E^{\text{pseudo}}\}$. 
Finally, the training data $\mathcal{D}$ is a combination of the small amount of human-annotated data,
and a large amount of data with pseudo labels and explanations.
We optimize our C-NAT with the fully-supervised training objective function on the combined training dataset.

\subsection{Unsupervised Learning}
\label{ssec:unsuper}
For the real-world scenario where no human annotated explanations are available, we also explore the unsupervised learning strategy for our C-NAT model training. 
Different from the autoregressive interpretation approach, golden explanations are only used as the training target but not the decoder input for the non-autoregressive interpretation approach.
To mimic the human-annotated training targets, we utilize the back-translation mechanism to generate pseudo explanations as the noisy training targets, and keep refining the explanation generation with a pre-trained language model during training.

\section{Experiments}

\begin{table}[t]
	\centering
\resizebox{0.5\textwidth}{!}{
	\begin{tabular}{l c c c c c c c}
		\toprule
		\textbf{Methods}
		& \textbf{Rationality}$^{\uparrow}$ 
		& \textbf{PPL}$^{\downarrow}$ & \textbf{NE-Acc}$^{\uparrow}$	& \textbf{Acc}$^{\uparrow}$ 
		& \textbf{Latency}$^{\downarrow}$ & \textbf{Speedup}$^{\uparrow}$
		\\ \hline
		Transformer(AT) &  76.51 & 31.74 & 81.27 & 84.11 & 566ms & $1.00\times$ \\
		\hline
		C-NAT &	\textbf{77.41} &	 30.61 & \textbf{85.01} & \textbf{87.14} & \textbf{34ms} & $\mathbf{16.65\times}$ \\
		\ \ \ w/o LM  & 75.79 & 44.31 & 85.01 & 86.95 & 34ms & $16.65\times$ \\
		\ \ \ w/o NAR & 71.78 & 47.24 & 85.01 & 84.13 & 518ms & $1.09\times$ \\
		\ \ \ w/o LCE & 76.09 & \textbf{30.24} & 42.65 & 47.15 & 34ms & $16.65\times$ \\
		\bottomrule
	\end{tabular}}
\caption{ Automatic evaluation results in the Spouse Prediction task with weak-supervison. The higher$^{\uparrow}$(or smaller$^{\downarrow}$), the better. At the bottom, we present the results of ablation study.}
\label{weak-sup}
\end{table}

\begin{table}[t]
	\centering
	\small
\resizebox{0.45\textwidth}{!}{
	\begin{tabular}{p{45pt} c c c c c c}
		\toprule
		\textbf{Methods}
		& \textbf{Rationality}$^{\uparrow}$ 
		& \textbf{PPL}$^{\downarrow}$ & \textbf{Inter-Rep}$^{\downarrow}$	& \textbf{Acc}$^{\uparrow}$ \\ \hline
		C-NAT &	\textbf{72.69} &	 46.32 & \textbf{0.49} & \textbf{83.12} \\
		\ \ \ w/o LM  & 70.49 & 57.46 & 0.52 & 83.00 \\
		\ \ \ w/o SSup & 4.68 & \textbf{37.26} & 0.89 & 82.36 \\
		\bottomrule
	\end{tabular}}
\caption{Automatic evaluation results in NLI task with the unsupervised learning strategy. 
The higher$^{\uparrow}$(or smaller$^{\downarrow}$), the better. At the bottom, we present the results of ablation study.
}
\label{unsup}
\end{table}

\subsection{Tasks and Datasets}
To verify the effectiveness of our approach, we conduct experiments on the Natural Language Inference (NLI) and Spouse Prediction (SP) tasks. NLI task aims to predict the entailment relationship between two sentences. SP task is to predict whether two people in the given sentence are spouses.

We use three datasets as our testbeds for \textbf{fully-supervised}, \textbf{weakly-supervised} and \textbf{unsupervised} learning respectively: \textbf{e-SNLI} \cite{camburu2018snli}, \textbf{SP} \cite{hancock2018training}, and \textbf{SNLI} \cite{bowman2015large}.
SNLI is a standard benchmark for the NLI task, while e-SNLI extends it with human-annotated natural language explanations for each sentence pair. Therefore, we use the e-SNLI dataset to generate explanations with full-supervision, while using the SNLI dataset for unsupervised explanation generation. SP dataset has only 30 samples annotated with human explanations, which we thus adopt for weakly-supervised explanation generation. 
Besides, as introduced in Section \ref{ssec:weak} and \ref{ssec:unsuper},  we propose two methods to generate pseudo data in low-resource scenarios. For the SP dataset, we extract templates from 30 human explanations, which are then used to generate pseudo explanations and form our \textbf{SP-Pseudo datase}t. For the SNLI dataset without human annotated explanations at all, we propose to use a pre-trained NMT model to generate pseudo explanations, which form our \textbf{SNLI-Pseudo dataset}. The statistics of all datasets and pseudo data are shown in Table \ref{data_stt}.

\subsection{Metrics}
To evaluate classification performance and explanation quality, we report  
\textbf{NE-Acc}(classification accuracy without generating explanations), \textbf{Acc} (classification accuracy), \textbf{BLEU} (similarity between generation and ground truth, if any), \textbf{PPL} (fluency of generated explanations), \textbf{Inter-Rep} (diversity of generated explanations), and \textbf{Rationality} (rationality of explanations).
Specifically, the Rationality metric is a model-based evaluation metric, which utilizes a pre-trained classifier to evaluate whether the generated explanation is reasonable for corresponding input and prediction.

\subsection{Implementation Details}
We set the embedding size and the hidden size as 512, and use 8 heads. The layer number of the encoder and decoder are set as both 6. 
We use the Adam \cite{kingma2014adam} for optimization with $\beta_1=0.9$, $\beta_2=0.999$ and $\epsilon=10^{-8}$. The learning rate is set to 0.00004, and the dropout rate is set to 0.3. 

\subsection{Results of Fully-Supervised Learning}
Evaluation results on e-SNLI in the full-supervised learning scenario are shown in Table \ref{result_nli_overall}. 
We observe that our C-NAT can achieve the comparable performance of explanation generation and label prediction with more than $20 \times$ speedup compared to the baseline autoregressive models.
We also conduct the ablation study to evaluate the effectiveness of each component. 
We find that the BLEU score and PPL score drop significantly with the LM discriminator removed, but the prediction accuracy remains.  It indicates that the pre-trained language model can effectively improve the fluency of generated explanations.
If we modify C-NAT for autoregressive generation, much higher inference latency would be witnessed, and the performance would also degrade due to the exposure bias problem.
Besides, we notice the classification performance drops on NE-Acc and ACC for baseline models, while our C-NAT achieves 2.07 absolute improvement. This demonstrates that our method can improve the inference ability of the classifier with model interpretability increased, instead of improving interpretability at the cost of classification performance.

\subsection{Results of Weakly-Supervised Learning}
Table \ref{weak-sup} shows the results of our C-NAT model with weakly-supervised learning strategy on the Spouse Prediction dataset that has only 30 human annotated explanations. We augment with pseudo data generated by our template-based approach.
Because there are no previous works exploring the weakly-supervised learning method for explanation generation, we choose the modified Transformer model supporting classification as our baseline.
Despite the small amount of human-annotated data, with the pseudo labels and explanations, we can still achieve improvement compared to the baseline model on all metrics.

\subsection{Results of Unsupervised Learning}
We conduct experiments in the Natural Language Inference task under the unsupervised learning scenario where no human-annotated explanations are available.
Table \ref{unsup} shows the experimental results of applying our approach in such a scenario. We observe that the LM clearly affects the persuasion accuracy and the fluency of explanations. Moreover, we also notice that the performance drops a lot without using the unsupervised learning strategy, which confirms the effectiveness of our unsupervised learning approach.

\section{Conclusion}
\label{sec:print}
In this paper, we explore the important problem of generating human-friendly sentence-level explanations in low-resource scenarios. To solve the high inference latency problem of previous interpretable models, we propose our C-NAT to support parallel explanation generation and simultaneous prediction.
We conduct extensive experiments in the Natural Language Inference task and Spouse Prediction task in the fully-supervised learning, weakly-supervised learning, and unsupervised learning scenarios. 
Experimental results reveal that our C-NAT can generate fluent and diverse explanations with classification performance also improved.

\vfill\pagebreak

\small
\bibliographystyle{IEEEbib}
\bibliography{refs}

\end{document}